\documentclass[9pt]{elife} 
\usepackage{bm}
\usepackage{multicol}
\usepackage{multirow}
\usepackage{wrapfig}

\usepackage[utf8]{inputenc} 
\usepackage[T1]{fontenc}    
\usepackage{hyperref}       
\usepackage{url}            
\usepackage{booktabs}       
\usepackage{amsmath}
\usepackage{nicefrac}       
\usepackage{microtype}      
\usepackage{hhline}
\usepackage{graphicx}
\usepackage[font=small,labelfont=bf,tableposition=top]{caption}
\DeclareCaptionLabelFormat{andtable}{#1~#2  \&  \tablename~\thetable}
\usepackage[english]{babel}
\usepackage[english]{babel}
\usepackage{float}
\usepackage{wrapfig}
\usepackage{amsthm}
\usepackage{listings}
\usepackage{courier}
\definecolor{codegreen}{rgb}{0,0.6,0}
\definecolor{codegray}{rgb}{0.5,0.5,0.5}
\definecolor{codepurple}{rgb}{0.58,0,0.82}
\definecolor{backcolour}{rgb}{0.95,0.95,0.92}

\usepackage{soul}
\sethlcolor{white}

\lstdefinestyle{mystyle}{
    float=tp,
    floatplacement=tbp,
    commentstyle=\color{codegreen},
    keywordstyle=\color{magenta},
    numberstyle=\tiny\color{codegray},
    stringstyle=\color{codepurple},
    basicstyle=\ttfamily\small,
    breakatwhitespace=false,         
    breaklines=true,                 
    captionpos=b,                    
    keepspaces=true,                 
    numbers=left,                    
    numbersep=5pt,                  
    showspaces=false,                
    showstringspaces=false,
    showtabs=false,                  
    tabsize=2,
    frame=tb,
}

\newfloat{lstfloat}{t}{lop}
\floatname{lstfloat}{Listing}

\title{EspalomaCharge: Machine learning-enabled ultra-fast partial charge assignment}
\author[1, 2]{Yuanqing~Wang~(ORCID:~\href{https://orcid.org/0000-0003-4403-2015}{0000-0003-4403-2015})}
\author[1]{Iván~Pulido~(ORCID:~\href{https://orcid.org/0000-0002-7178-8136}{0000-0002-7178-8136})}
\author[1, 3]{Kenichiro~Takaba~(ORCID:~\href{https://orcid.org/0000-0002-2481-8830}{0000-0002-2481-8830})}
\author[1, 4]{Benjamin~Kaminow~(ORCID:~\href{http://orcid.org/0000-0002-2266-3353}{0000-0002-2266-3353})}
\author[1]{Jenke~Scheen~(ORCID:~\href{http://orcid.org/0000-0001-9781-0445}{0000-0001-9781-0445})}
\author[1, 5]{Lily~Wang~(ORCID:~\href{https://orcid.org/0000-0002-6095-6704}{0000-0002-6095-6704})}
\author[1]{John~D.~Chodera~(ORCID:~\href{http://orcid.org/0000-0003-0542-119X}{0000-0003-0542-119X})}

\affil[1]{Computational and Systems Biology Program, Sloan Kettering Institute, Memorial Sloan Kettering Cancer Center, New York, NY 10065}
\affil[2]{Physiology, Biophysics, and System Biology Ph.D.\ Program, Weill Cornell Medical College, Cornell University, New York, NY 10065}
\affil[3]{Pharmaceutical Research Center, Advanced Drug Discovery, Asahi Kasei Pharma Corporation, Shizuoka 410-2321, Japan}
\affil[4]{Tri-Institutional PhD Program in Computational Biology and Medicine, Weill Cornell Medical College, Cornell University,
New York, NY 10065}
\affil[5]{Open Molecular Sciences Foundation, Davis, CA 95618}

\corr{yuanqing.wang@choderalab.org}{YW}
\corr{john.chodera@choderalab.org}{JDC}
\begin{document}

\maketitle

\begin{abstract}
Atomic partial charges are crucial parameters in molecular dynamics (MD) simulation, dictating the electrostatic contributions to intermolecular energies, and thereby the potential energy landscape.
Traditionally, the assignment of partial charges has relied on surrogates of \textit{ab initio} semiempirical quantum chemical methods such as AM1-BCC, and is expensive for large systems or large numbers of molecules.
We propose a hybrid physical / graph neural network-based approximation to the widely popular AM1-BCC charge model that is orders of magnitude faster while maintaining accuracy comparable to differences in AM1-BCC implementations.
Our hybrid approach couples a graph neural network to a streamlined charge equilibration approach in order to predict molecule-specific atomic electronegativity and hardness parameters, followed by analytical determination of optimal charge-equilibrated parameters that preserves total molecular charge.
This hybrid approach scales linearly with the number of atoms, enabling, for the first time, the use of fully consistent charge models for small molecules and biopolymers for the construction of next-generation self-consistent biomolecular force fields.
Implemented in the free and open source package \texttt{espaloma\_charge}, this approach provides drop-in replacements for both AmberTools \texttt{antechamber} and the Open Force Field Toolkit charging workflows, in addition to stand-alone charge generation interfaces.
Source code is available at \url{https://github.com/choderalab/espaloma_charge}.
\end{abstract}


Molecular mechanics (MM) force fields abstract atoms as point charge-carrying particles, with their electrostatic energy ($U_e$) calculated by some Coulomb's law~\cite{coulomb1785premier}
\begin{equation}
U_e(r_{ij}) = k_e \frac{q_i q_j}{r_{ij}},
\end{equation}
(or some modified form), where $k_e$ is Coulomb constant (energy * distance$^2$ / charge$^2$) and $r_{ij}$ the interatomic distance.
In fixed-charge molecular mechanics force fields, the partial charges $q_i$ are
treated as constant, static parameters, agnostic of instantaneous geometry.
As such, partial charge assignment---the manner in which partial charges are assigned to each atom in a given system based on their chemical environments---plays a crucial role in molecular dynamics (MD) simulation, determining the electrostatic energy ($U_e$) at every step and shaping the energy landscape.

\begin{figure}
    \centering
    \includegraphics[width=0.8\textwidth]{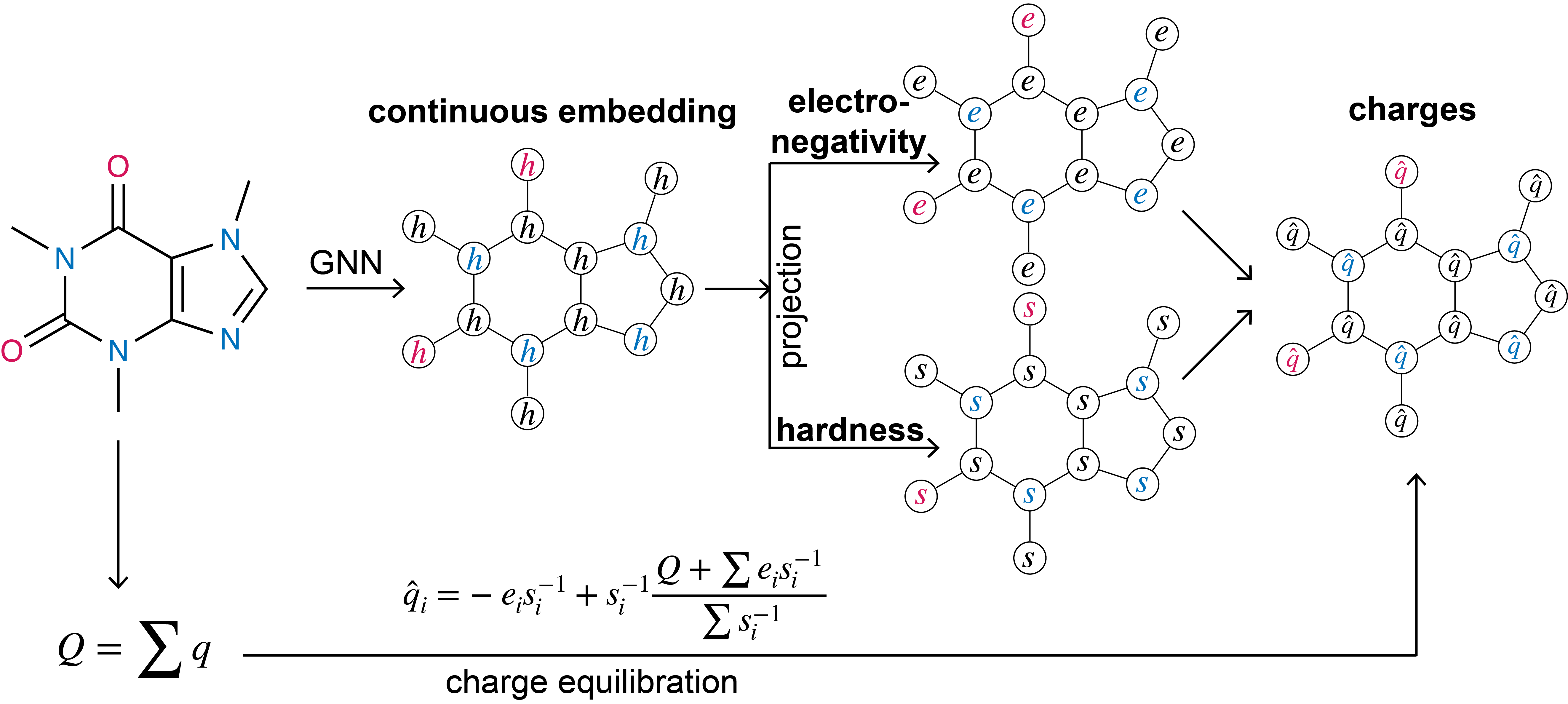}
    \caption{
    \textbf{Schematic overview of EspalomaCharge: a hybrid physical / GNN model for fast charge assignment.}
    First, the graph node representation $h$ assigned by a GNN is used to compute unconstrained electronegativity $e_i$ and hardness $s_i$ to each atom.
    Second, the charge potential energy is minimized analytically to yield predicted partial charges $\hat{q}_i$ that satisfy the total molecular charge constraint $Q$.
    }
    \label{fig:abstract}
\end{figure}

\paragraph{Traditionally, partial charges have been derived from expensive \textit{ab initio} or semi-empirical quantum chemical approaches}
In the early stages of development of molecule mechanics (MM) force fields, \textit{ab initio} methods were used to generate electrostatic potentials on molecular surfaces from which restrained electrostatic potential (RESP) charge fits were derived~\cite{doi:10.1021/j100142a004}.
This process proved to be expensive, especially for large molecules or large numbers of molecules (e.g., in virtual screening, where datasets now approach $10^9$ molecules~\cite{glaser2021high}).
This led to the development of the AM1-BCC charge scheme~\cite{jakalian2000fast,jakalian2002fast}---a method for \textit{approximating} restrained electrostatic potential (RESP) fits at the HF/6-31G* level of theory, by first calculating population charges using the much less expensive AM1 semiempirical level of theory and subsequently correcting charges via \emph{bond charge corrections (BCC)}.
As a result, this approach has been widely adopted by the molecular mechanics community utilizing force fields such as GAFF~\cite{doi:10.1002/jcc.20035} and the Open Force Field force fields~\cite{qiu2021development}.

Despite this progress, there are still multiple drawbacks with AM1-BCC.
First, the computation is dependent on the generation of one or more conformers, which contributes to the discrepancy among the results of different chemoinformatics toolkits.
While conformer ensemble selection methods such as ELF10~\footnote{
ELF10 denotes that the ELF ("electrostatially least-interacting functional groups") conformer selection process was used to generate 10 diverse conformations from the lowest energy 2\% of conformers. 
Electrostatic energies are assessed by computing the sum of all Coulomb interactions in vacuum using the absolute values of MMFF charges assigned to each atom~\cite{mmff}. 
AM1-BCC charges are generated for each conformer and then averaged.} attempt to minimize these geometry-dependent effects, they do not fully eliminate them, and significant discrepancies between toolkits can remain.

Secondly, the speed is still a bottleneck (especially when it comes to the virtual screening of large libraries) as it still requires QM calculation for the parameterization.
Moreover, the runtime complexity of AM1-BCC scales $\mathcal{O}(N^2)$ in the number of atoms $N$.
In particular, the poor runtime complexity necessitates using a different charging model for biopolymers (such as proteins and nucleic acids), making the process of extending these polymeric force fields to accommodate post-translational modifications, nonstandard residues, covalent ligands, and other chemical modifications both complex and likely to require a \emph{third} charging strategy within the same simulation.

\paragraph{Machine learning approaches to charge assignment have recently been proposed but face challenges in balancing generalization with the ability to preserve total molecular charge}
The rising popularity of machine learning has led to a desire to exploit new approaches to rapidly predict partial atomic charges.
For example, recent work from \citet{doi:10.1021/acs.jcim.7b00663} employed a random forest approach to assign charges based on atomic features, but faced the issue of needing to preserve total molecular charge while making predictions on an atomic basis---they distribute the difference between predicted and reference charge evenly among atoms.
Similarly, \citet{metcalf2020electron} preserves the total charge by allowing only charge transfer in message-passing form resulting in zero net-charge change.
A more classical approach by \citet{doi:10.1021/ci034148o} tackles the charge constraint problem in a clever manner: instead of directly predicting charges, by predicting atomic electronegativity and electronic hardness, a simple constrained optimization problem inspired by physical charge equilibration~\cite{doi:10.1021/j100161a070} can be solved analytically to yield partial charges that satisfy total molecular charge constraints. 
In spite of its experimental success, its ability to reproduce quantum chemistry-based charges is heavily dependent upon the discrete atom typing 
scheme to classify and group atoms by their chemical environments.

Recently, \citet{D2SC02739A} designed a graph neural networks-based atom typing scheme, termed \textbf{Espaloma} (extensible surrogate potential optimized by message-passing algorithms), to replace the human expert-derived, discrete atom types with continuous atom embeddings (Figure~\ref{fig:abstract}).
This allows atoms with subtle chemical environment differences to be distinguished by the model, without the need to painstakingly specify heuristics.
\begin{lstlisting}[language=Python, label=lst:install, numbers=none, caption=\textbf{Installing EspalomaCharge via the pip Python package manager.}]
$ pip install espaloma_charge
\end{lstlisting}
\begin{lstlisting}[language=Python, label=lst:use, numbers=none, caption={\textbf{Example illustrating the EspalomaCharge Python API.} Here, EspalomaCharge assigns AM1-BCC ELF10 equivalent partial charges to an RDKit Molecule, returning them in a NumPy array.}]
>>> from rdkit import Chem; from espaloma_charge import charge
>>> molecule = Chem.AddHs(Chem.MolFromSmiles("CCO"))
>>> charge(molecule)
array([ 0.95081186, -1.007628  ,  0.93298626, -0.1128267 , -0.1128267 ,
       -0.1128267 , -0.10835528, -0.10835528, -0.32097816], dtype=float32)
\end{lstlisting}

\paragraph{EspalomaCharge generates AM1-BCC ELF10 quality charges in an ultra-fast manner using machine learning}

In this paper, we use the continuous embedding atom representation scheme from \textbf{Espaloma} in conjunction with analytical constrained charge assignment inspired by charge equilibration to come up with an ultra-fast machine learning surrogate for partial charge assignment (EspalomaCharge).
We train EspalomaCharge on an expanded set of protonation states and tautomers of representative biomolecules and druglike molecules (the SPICE dataset~\cite{https://doi.org/10.48550/arxiv.2209.10702}) to assign high-quality AM1-BCC ELF10 charges~\cite{jakalian2002fast}.
The resulting EspalomaCharge model accurately reproduces AM1-BCC ELF10 charges to an error well within the discrepancy between AmberTools \texttt{sqm} and OpenEye \texttt{oequacpac} implementations on average 2,000 times faster than AmberTools on the SPICE dataset, can utilize either CPU or GPU, and scales as $\mathcal{O}(N)$ with number of atoms, allowing even entire proteins to be assigned AM1-BCC equivalent charges.
We implement this approach in the Python package \texttt{espaloma\_charge}, which is distributed open source under MIT license and \texttt{pip}-installable (Listing~\ref{lst:install}).

\section*{Theory: Espaloma graph neural networks for chemical environment perception, charge equilibration (QEq), and EspalomaCharge}
\label{sec:theory}

\paragraph{Espaloma uses graph neural networks to perceive atomic chemical environments}

\textbf{Espaloma}~\cite{D2SC02739A} uses \textit{graph neural networks (GNNs)}~\cite{DBLP:journals/corr/KipfW16, xu2018powerful, gilmer2017neural, hamilton2017inductive, battaglia2018relational, https://doi.org/10.48550/arxiv.2301.08893} to assign continuous latent representations of chemical environments to atoms that replace human expert-derived discrete atom types.
These continuous atom representations are subsequently used to assign symmetry-preserving parameters for atomic, bond, angle, torsion, and improper force terms.

When GNNs are employed in chemical modeling, the atoms are abstracted as nodes ($v$) and bonds as edges ($e$) of a graph $\mathcal{G}$.
$h_v^{(0)}$, the initial features associated with node $v$ are determined based on resonance-independent atomic chemical features from a cheminformatics toolkit (see Section~\ref{sec:detailed}).
Following the framework from \citet{gilmer2017neural, xu2018powerful, battaglia2018relational}, for a node $v$ with neighbors $u \in \mathcal{N}(v)$, in a graph $\mathcal{G}$, with $h_v^{(k)}$ denoting the feature of node $v$ at the $k$-th layer (or $k$-th round of message-passing) and $h_v^{0} \in \mathbb{R}^{C}$ the initial node feature on the embedding space, the $k$-th message-passing step of a GNN can be written as three steps:
First, an \textit{edge update},
\begin{equation}
\label{eq:edge_update}
h_{e_{uv}}^{(k+1)} = \phi^{e}
\big(
h_u^{(k)}, h_v^{(k)}, h_{e_{uv}}^{(k)}
\big),
\end{equation}
where the feature embeddings $h_u$ of two connected nodes $u$ and $v$ update their edge feature embedding $h_{e_{uv}}$,
followed by \textit{neighborhood aggregation},
\begin{equation}
\label{eq:neighborhood_aggregation}
a_v^{(k+1)} = \rho^{e \rightarrow v} (\{ h_{e_{uv}}^{(k)}, u \in \mathcal{N}(v) \}),
\end{equation}
where edges incident to a node $v$ pool their embeddings to form \textit{aggregated neighbor embedding} $a_v$,
and finally a \textit{node update},
\begin{equation}
\label{eq:node_update}
h_v^{(k+1)} = \phi^{v}(a_v^{(k+1)}, h_v^{(k)})
\end{equation}
where $\mathcal{N}(\cdot)$ denotes the operation to return the multiset of neighbors of a node and $\phi^e$ and $\phi^v$ are implemented as feed-forward neural networks.
Since the neighborhood aggregation functions $\rho^{e \rightarrow v}$ are always chosen to be indexing-invariant functions, namely \texttt{SUM} or \texttt{MEAN} operator, Equation~\ref{eq:neighborhood_aggregation}, and thereby the entire scheme, is permutationally invariant.
In practice, choices such as dimensionality of node and edge vectors, number of layers, layer width, activation function, aggregation operators, and initial conditions for training are treated as hyperparameters and optimized during training to produce robust, near-optimal models on a held-out validation set separate from a test set.

\paragraph{Charge equilibration (QEq) is a physically inspired model for computing partial charges while maintaining total molecular charge}

This \textbf{Espaloma} framework can be used to predict atomic parameters that can be fed into subsequent neural modules that predict molecular mechanics parameters. 
For partial charges, however, the constraint that the predicted partial charges $\hat{q}_i$ should sum up to the total charge $Q$---the sum of all formal charges or total molecular charge---is non-trivial to satisfy were the charges to be predicted directly.
\begin{equation}
\sum \hat{q}_i = \sum q_i = Q.
\end{equation}
We adopt the method proposed by \citet{doi:10.1021/ci034148o} where we predict the \textit{electronegativity} $e_i$ and \textit{hardness} $s_i$ of each atom $i$, which are defined as the first- and second-order derivative of the potential energy in charge equilibration approaches~\cite{doi:10.1021/j100161a070}:
\begin{equation}
e_i \equiv \frac{\partial U_e}{\partial q_i};
s_i \equiv \frac{\partial^2 U_e}{\partial q_i^2}.
\end{equation}
Next, we minimize the second-order Taylor expansion of the charging potential energy contributed by these terms, neglecting interatomic electrostatic interactions:
\begin{equation}
\{ \hat{q}_i \} = \operatorname{argmin}\limits_{q_i}
\sum\limits_i (\hat{e}_i \hat{q_i} + \frac{1}{2} \hat{s}_i \hat{q}_i^2),
\end{equation}
which, as it turns out, has an analytical solution given by Lagrange multipliers:
\begin{equation}
\label{eq:charge-equilibration-solution}
\hat{q}_i = - e_i s_i^{-1} + s_i^{-1} \frac{Q + \sum e_i s_i^{-1}}{\sum s_i^{-1}}.
\end{equation}
We thus use the Espaloma framework to predict the unconstrained atomic electronegativity ($e$) and hardness ($s$) parameters used in Equation~\ref{eq:charge-equilibration-solution} to assign partial charges in a manner that ensures total molecular charge sums to $Q$.
It is worth noting that, by the equivalence analysis proposed in \citet{D2SC02739A}, the tabulated atom typing scheme \citet{doi:10.1021/ci034148o} uses amounts to a model working analogously to a Weisfeiler-Lehman test~\cite{weisfeiler1968reduction} with hand-written kernel, whereas here we replace this with an end-to-end differentiable GNN model to greatly expand its resolution and ability to optimize based on reference charges.

\paragraph{EspalomaCharge has $\mathcal{O}(N)$ time complexity in the number of atoms}

One of the primary advantages of spatial GNNs that pass messages among local neighborhoods is their $\mathcal{O}(E)$ complexity, where $E$ is the number of edges.
In chemical modeling, since the sparsity of the graph is roughly fixed (number of edges is $3$ to $4$ times that of number of nodes), it is safe to write the runtime complexity as $\mathcal{O}(N)$, with $N$ being the number of nodes (atoms).
The charge equilibration (QEq) step with its linear operator does not alter the complexity, nor is it the bottleneck of EspalomaCharge.
Therefore, unlike with \textit{ab initio} or semi-empirical methods, the runtime complexity of EspalomaCharge is $\mathcal{O}(N)$.

\section*{Experiments: EspalomaCharge accurately reproduces AM1-BCC charges at a fraction of its cost}



We show, in this section, that the discrepancy between EspalomaCharge and the OpenEye toolkit is comparable to or smaller than that between AmberTools~\cite{amber2020} and OpenEye.
EspalomaCharge is fast and scalable to larger systems, taking seconds to parameterize a biopolymer with 100 residues on CPU.

\paragraph{The SPICE dataset covers biochemically and biophysically interesting chemical space}

To curate a dataset representing the chemical space of interest for biophysical modeling of biomolecules and drug-like small molecules, we use the SPICE~\cite{https://doi.org/10.48550/arxiv.2209.10702} dataset, enumerating reasonable protonation and tautomeric states with the OpenEye Toolkit.
We generated AM1-BCC ELF10 charges for each of these molecules using the OpenEye Toolkit, and trained EspalomaCharge ({\bf Figure~\ref{fig:abstract}}) to reproduce the partial atomic charges with a squared loss function.
This model, with its parameters distributed with the code, is used in all characterization results hereafter.

\begin{table}[htbp]
    \centering
    \resizebox{\textwidth}{!}{%
    \begin{tabular}{c c c c c c c c c}
    \hline
    \multirow{2}{*}{dataset}
    & \multirow{2}{*}{$N_\text{mol}$}
    & \multirow{2}{*}{avg.~$N_\text{atoms}$}
    & \multicolumn{2}{c}{average $\mathtt{RMSE}$ (e)}
    & \multicolumn{3}{c}{average walltime (s)} \\

    & & & \footnotesize{|EspalomaCharge - OpenEye|} & \footnotesize{|AmberTools - OpenEye|}
    & EspalomaCharge & AmberTools & OpenEye\\
    \hline
    \footnotesize{SPICE~\cite{https://doi.org/10.48550/arxiv.2209.10702} test set} 
    & $29079$
    & $39.36$
    & $\mathbf{0.0435}_{0.0432}^{0.0438}$
    & $0.0623_{0.0618}^{0.0628}$
    & $\mathbf{0.05}$ & $93.10$ & $3.79$
    \\
    \hline
     \footnotesize{FDA approved} 
     & $1019$
     & $34.80$
     & $\mathbf{0.0266}_{0.0255}^{0.0281}$
     & $\mathbf{0.0244}_{0.0227}^{0.0263}$
     & $\mathbf{0.03}$
     & $46.15$
     & $1.87$
     \\
     \footnotesize{ZINC250K~\cite{doi:10.1021/acscentsci.7b00572}} 
     & $220250$
     & $42.70$
     & $\mathbf{0.0187}_{0.0187}^{0.0187}$
     & $0.0197_{0.0197}^{0.0198}$
     & $\mathbf{0.05}$
     & $124.89$
     & $3.63$
     
     \\
     \footnotesize{FreeSolv~\cite{duarte2017approaches}} 
     & $641$
     & $18.10$
     & $0.0110_{0.0104}^{0.0117}$
     & $\mathbf{0.0067}_{0.0057}^{0.0077}$
     & $\mathbf{0.03}$
     & $9.62$
     & $0.43$
     \\

     \footnotesize{PDB expo~\cite{10.1093/nar/28.1.235}}
     & $23399$
     & $35.94$
     & $\mathbf{0.0186}_{0.0184}^{0.0188}$
     & $0.0232_{0.0229}^{0.0236}$
     & $\mathbf{0.04}$
     & $88.86$
     & $3.63$ \\
     \hline
    \end{tabular}}
    \caption{
        \textbf{
        EspalomaCharge accurately and efficiently reproduces AM1-BCC charges for a wide variety of chemical spaces.
        }
        Here, $N_\text{mol}$ denotes the number of molecules in the dataset;
        avg.~$N_\text{atoms}$ denotes the average number of atoms in molecules for the corresponding dataset;
        average~$\mathtt{RMSE}$ is the charge RMS deviation between AM1-BCC implementations averaged over all molecules in the dataset, with sub- and superscripts denoting the 95\%-confidence interval of the mean (computed by bootstrapping over molecules in the dataset with replacement);
        average walltime denotes the average wall time for the respective toolkit to assign partial charges for a molecule in the dataset.
        Boldface statistics denote the best (most accurate or fastest) model or models (in case confidence intervals are indistinguishable) for each statistic.
    }
    \label{tab:in_n_out}
\end{table}

\begin{figure}
    \centering
    \includegraphics[width=1.0\textwidth]{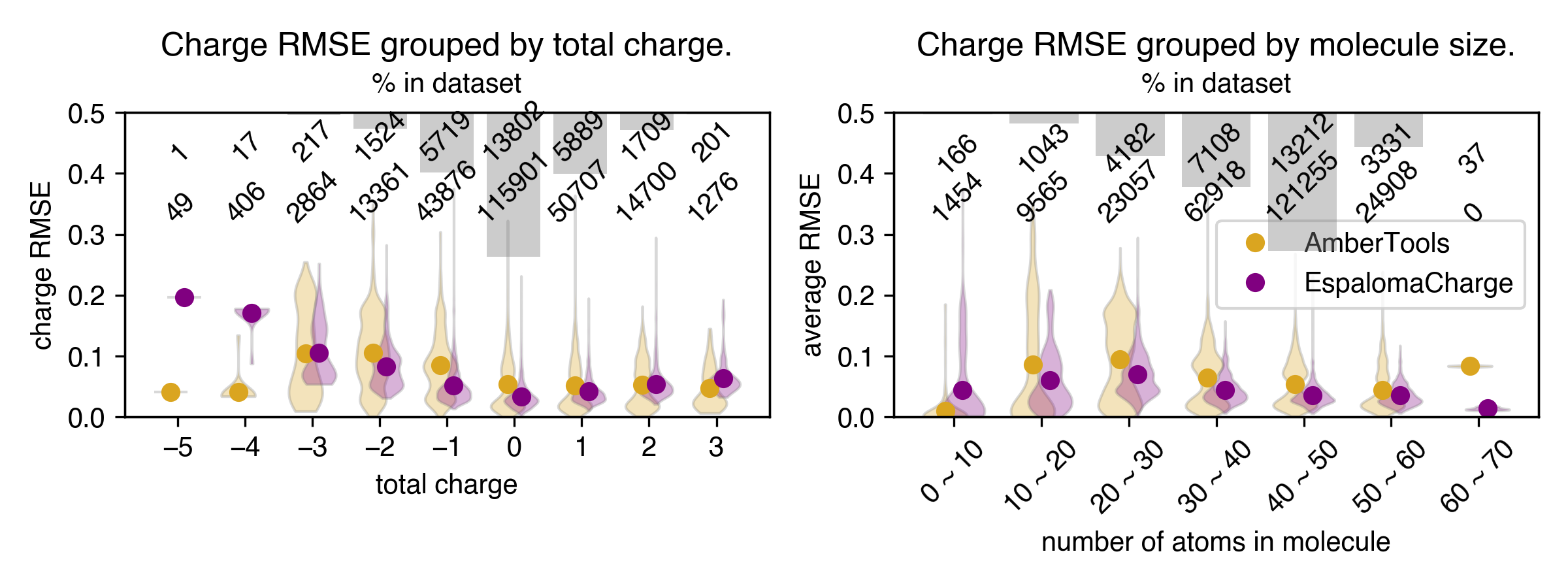}
    \caption{
    \textbf{EspalomaCharge shows smaller average charge RMSE than AmberTools on well-represented regions of chemical space.} 
    SPICE dataset test set performance stratified by total charge (\emph{left panel}) and molecule size (\emph{right panel}).
    To better illustrate the effects of limited training data on stratified performance, the number of test (upper number) and training (lower number) molecules falling into respective categories are also annotated with test set distribution plotted as histogram.
    }
    \label{fig:rmse_breakdown}
\end{figure}

\paragraph{EspalomaCharge is accurate, especially on chemical spaces where training data is abundant}

First, upon training on the 80\% training set of SPICE, we test on the 10\% held-out test set to benchmark the in-distribution (similar chemical specie) performance of EspalomaCharge (Table~\ref{tab:in_n_out}, first half).
Notably, the discrepancy (measured by charge RMSE) between EspalomaCharge and OpenEye is comparable with or smaller than that between AmberTools~\cite{amber2020} and OpenEye---two popular chemoinformatics toolkits for assigning AM1-BCC charges to small molecules.
Since it is a common practice in the community to use these two toolkits essentially \textit{interchangeably}, we argue that the discrepancy between these could be established as a baseline below which the error is no longer meaningful. 

We prepare several out-of-distribution external datasets to test the generalizability of EspalomaCharge to other molecules of significance to chemical and biophysical modeling, including a filtered list of FDA approved drugs, a subset of the ZINC~\cite{doi:10.1021/acscentsci.7b00572, pmid15667143} purchasable chemical space, and finally the FreeSolv~\cite{mobley2018open} dataset consisting of molecules with experimental and computationally-estimated solvation free energy.
The discrepancy between EspalomaCharge and OpenEye is lower than, or comparable with, that between AmberTools and OpenEye, demonstrating that the high performance of EspalomaCharge is generalizable, at least within chemical spaces frequently used in chemical modeling and drug discovery.

To pinpoint the source of the error for EspalomaCharge, we stratify the molecules by number of atoms and total molecular charge, computing the errors on each subset (Figure~\ref{fig:rmse_breakdown}).
Compared to the error baseline, EspalomaCharge is most accurate where there was abundant data in the training set.
This is especially true when it comes to stratification by net molecular charge, since the extrapolation from small systems to larger systems is encoded in the inductive biases of GNNs.
Given the performance of well-sampled charge bins, it seems likely the poor performance for molecules with more exotic $-4$ and $-5$ net charges will be resolved once the dataset is enriched with more examples of these states.

\begin{lstlisting}[language=Python, label=lst:api-openff, numbers=none, caption=\textbf{Example illustrating EspalomaCharge integration with the Open Force Field Toolkit.} Here\, EspalomaCharge is used to provide charges via the ToolkitWrapper facility.]
>>> from openff.toolkit.topology import Molecule
>>> from espaloma_charge.openff_wrapper import EspalomaChargeToolkitWrapper
>>> toolkit_registry = EspalomaChargeToolkitWrapper()
>>> molecule = Molecule.from_smiles("CCO")
>>> molecule.assign_partial_charges('espaloma-am1bcc', toolkit_registry=toolkit_registry)
>>> molecule.partial_charges
<Quantity([ 0.95081172 -1.00762811  0.93298611 -0.11282685 -0.11282685 -0.11282685
 -0.10835543 -0.10835543 -0.32097831], 'elementary_charge')>
\end{lstlisting}

\begin{lstlisting}[language=bash, label=lst:api-amber, numbers=none, caption=\textbf{Example illustrating the use of EspalomaCharge as a fast drop-in replacement for \texttt{sqm} in an AmberTools \texttt{antechamber} workflow.} By adding a single line\, EspalomaCharge can replace the slow \texttt{sqm}-based AM1-BCC model to provide fast charges for AmberTools based worklows.]
$ espaloma_charge -i in.mol2 -o in.crg
$ antechamber -fi mol2 -fo mol2 -i in.mol2 -o out.mol2 -c rc -cf in.crg 
\end{lstlisting}

It is worth mentioning that unified application programming interfaces (API) (See Listing~\ref{lst:api-openff}) integrated in Open Force Field toolkits are responsible for generating the performance benchmark experiments above.
Additionally, a command-line interface (CLI) is also provided for seamless integration of EspalomaCharge into Amber workflows (See Listing~\ref{lst:api-amber}).

\begin{figure}
    \centering
    \includegraphics[width=0.8\textwidth]{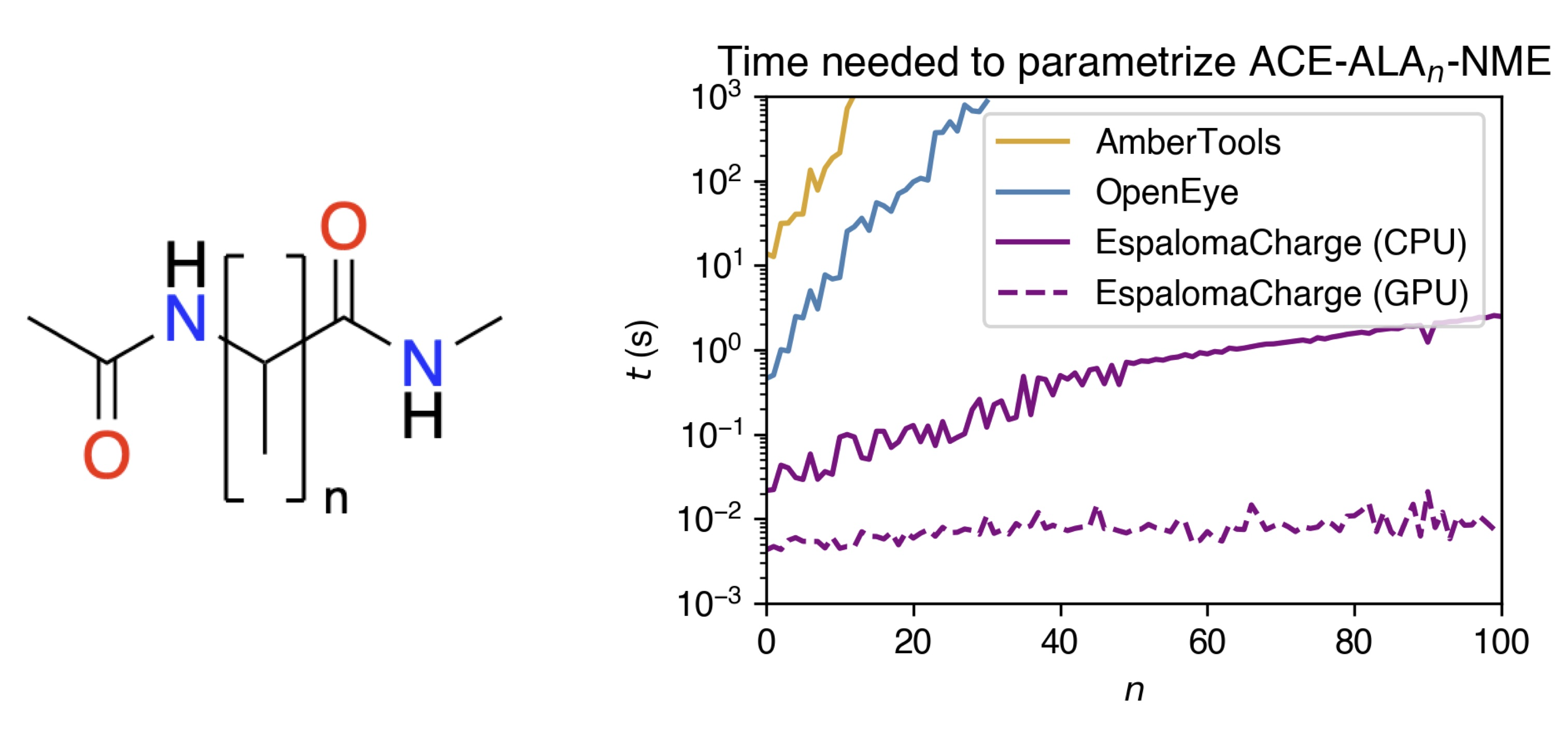}
    \caption{
        \textbf{EspalomaCharge is fast, even for large systems. }
        Wall time required to assign charges to ACE-ALA$_n$-NME peptides with different toolkits is shown on a log plot, illustrating that EspalomaCharge on the CPU or GPU is orders of magnitude faster than semiempirical-based charging methods for larger molecules or biopolymers, and is practical even for assigning charges to proteins of practical size.
        Fluctuation in traces is due to the stochasticity in timing trials.
    }
    \label{fig:alan}
\end{figure}


\paragraph{EspalomaCharge is fast, even on large biomolecular systems}

Apart from the accurate performance, the drastic difference in the speed of parameterization is also observed in the benchmarking experiments.
For the small molecule datasets in Table~\ref{tab:in_n_out}, EspalomaCharge is $300$ to $3000$ times faster than AmberTools and $15$ to $75$ times faster than OpenEye.

We closely examine the dependence of parameterization time on the size of the (biopolymer) system in Figure~\ref{fig:alan}, where we choose the peptide system ACE-ALA$n$-NME while varying $n=1,...,100$.
The parameterization wall time for AmberTools and OpenEye rapidly increase w.r.t. the size of the system (the theoretical runtime complexity for semi-empirical methods are $\mathcal{O}(N^2)$) and exceeds 1000 seconds at $n=18$ and $n=30$, respectively.
This scenario explains the infeasibility of employing AM1-BCC charges in parameterizing large systems.
EspalomaCharge, on the other hand, has $\mathcal{O}(N)$ complexity and is capable of parameterizing peptides of a few hundred residues within seconds.
This process can be further accelerated by distributing calculations on GPU hardware.

\begin{lstlisting}[language=Python, label=lst:batch, numbers=none, caption=\textbf{EspalomaCharge can be used in batch mode to rapidly charge virtual libraries.}]
>>> import timeit; from rdkit import Chem; from espaloma_charge import charge
>>> molecules = [Chem.MolFromSmiles("C" * idx) for idx in range(1, 101)] # linear alkanes
>>> [Chem.AddHs(molecule) for molecule in molecules]
>>> timeit.timeit(lambda: charge(molecules)) # single trial using CPU
2.9034676551818848
\end{lstlisting}
Batching many molecules into a single charging calculation can provide significant speed benefits when parameterizing large virtual libraries by making maximum use of hardware parallelism.
EspalomaCharge provides a seamless way to achieve these speedups when providing a \texttt{Sequence} of molecules, rather than single molecules at a time, as the input to the \texttt{charge} function in the API (Listing~\ref{lst:batch}).
In this case, the molecular graphs are \textit{batched} with their adjacency matrix concatenated diagonally, processed by GNN and QEq models, and subsequently \textit{unbatched} to yield the result.
For instance, the wall time needed to parameterize all 100 ACE-ALA$_n$-NME molecules from $n = 1,\ldots,100$ depicted in Figure~\ref{fig:alan} at once, in batch mode, is 7.11 seconds with CPU---only marginally longer than the time required to parameterize the largest molecule in the dataset, indicating that hardware resources are barely being saturated at this point.

\paragraph{Error from experiment in explicit solvent hydration free energies is not statistically significantly different between EspalomaCharge, AmberTools, and OpenEye implemnetations of AM1-BCC.}

\begin{figure}
    \centering
    \includegraphics[width=\textwidth]{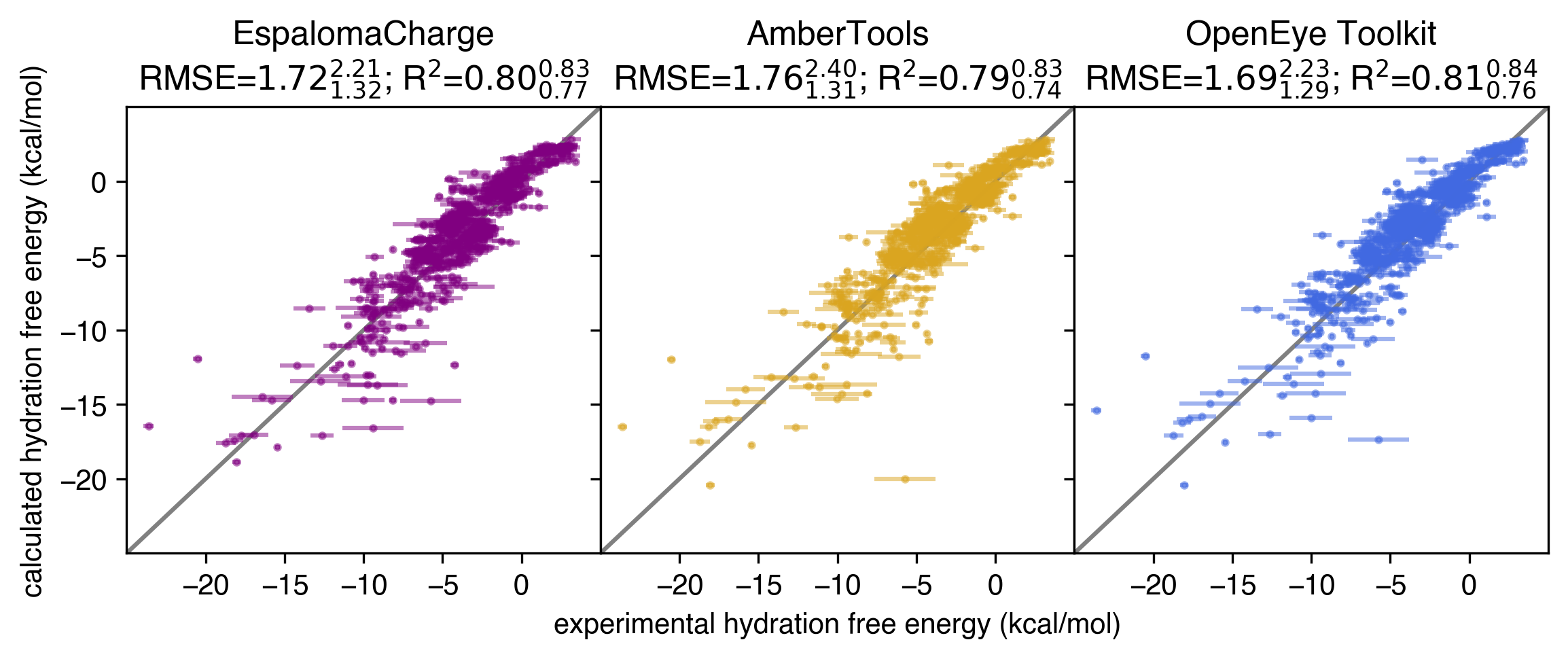}
    \caption{
        \textbf{EspalomaCharge introduces little error to explicit hydation free energy prediction.}
        Calculated-vs-experimental explicit solvent hydration free energies computed with AM1-BCC charges provided by EspalomaCharge, AmberTools, and the OpenEye Toolkit, respectively.
        Simulations used the GAFF 2.11 small molecule force field~\cite{doi:10.1002/jcc.20035} and TIP3P water~\cite{jorgensen1983comparison} with particle mesh Ewald electrostatics (see Detailed Methods).
        Annotated are root mean square error (RMSE) and R$^2$ score therebetween and bootstrapped 95\% confidence interval.
        See also Appendix Figure~\ref{fig:hydration_in} for comparison among computed hydration free energies. 
    }
    \label{fig:hydration}
\end{figure}

While the charge deviations between EspalomaCharge and other toolkit implementations of AM1-BCC are comparable to the deviation \emph{between} toolkits, it is unclear how the magnitude of these charge deviations translates into deviations of observable condensed-phase properties (such as free energies) from experiment.
To assess this, we carried out explicit solvent hydration free energy calculations, which serve as an excellent gauge of the impact of parameter perturbations~\cite{mobley2007comparison}, as the result is heavily dependent upon the small molecule charges.
We use each set of charges in calculating the hydration free energies for the molecules in FreeSolv~\cite{duarte2017approaches} (see Detailed Methods in Appendix Section~\ref{sec:detailed}), a standard curated dataset of experimental hydration free energies.
In Figure~\ref{fig:hydration}, we compare the computed explicit solvent hydration free energies with experimental measurements and quantify the impact of charge model on both deviation statistics (RMSE) and correlation statistics (R$^2$) with experiment.
We note that EspalomaCharge provides statistically indistinguishable performance compared to AmberTools~\cite{amber2020} and the OpenEye toolkit on both metrics, RMSE and R$^2$.
This encouraging result suggests that any discrepancy introduced by EspalomaCharge is unlikely to significantly alter the qualitative behavior of MD simulations in terms of ensemble averages or free energies.

\section*{Discussion}

\paragraph{EspalomaCharge assigns high-quality conformation-independent AM1-BCC charges using a modern machine learning infrastructure that supports accelerated hardware}

Composing the Espaloma graph neural networks framework~\cite{D2SC02739A, https://doi.org/10.48550/arxiv.1909.07903} for producing continuous, vectorial representations of the chemical environment of individual atoms with a conformation-independent charge equilibration (QEq) scheme~\cite{doi:10.1021/ci034148o} for assigning partial atomic charges that satisfy total molecular charge constraints, EspalomaCharge provides a robust approach for assigning conformer-agnostic AM1-BCC charges to biomolecular systems.
Because EspalomaCharge is built on PyTorch~\cite{paszke2017automatic}, a fast, modern, Python-based machine learning framework, it supports multiple optimized compute backends, including both CPUs and GPUs. 
Unlike AM1-BCC implementations based on traditional semiempirical quantum chemical codes, EspalomaCharge has $\mathcal{O}(N)$ runtime complexity with respect to the number of atoms $N$ (Figure~\ref{fig:alan}), and introduces only small discrepancies to high-quality AM1-BCC reference implementations comparable to the discrepancies among popular AM1-BCC implementations (Table~\ref{tab:in_n_out}).

\paragraph{The ability to assign topology-driven conformation-independent self-consistent charges to small molecules and biopolymers prepares the community for next-generation unified force fields}

EspalomaCharge, thanks to its $\mathcal{O}(N)$ runtime complexity, can assign charges to biopolymers with hundreds of residues---including proteins with exotic post-translational modifications or covalent ligands, nucleic acids, or complex conjugates of multiple kinds---within seconds.
For the first time, rather than using multiple distinct methodologies to parameterize various components in a system (e.g., RESP-derived charges for amino acids and AM1-BCC charges for noncovalent ligands), it is feasible to simultaneously and self-consistently parameterize small molecules and biopolymers (and more complex covalent modifications of biopolymers) with a high-quality self-consistent scheme.
This would be compatible with the next generation of unified force fields for small molecules and biopolymers, namely \citet{D2SC02739A}.

\paragraph{EspalomaCharge provides a simple API and CLI for facile integration into popular workflows}

EspalomaCharge is a \texttt{pip}-installable (Listing~\ref{lst:install}) open software package (Appendix Section~\ref{sec:detailed}), making it easy to integrate into existing workflows with minimal complexity.
Assigning charges to molecules using the EspalomaCharge Python API is simple and streamlined (Listing~\ref{lst:use}).
A GPU can be used automatically, and entire libraries can be rapidly parameterized in batch mode (Listing~\ref{lst:batch}).
EspalomaCharge provides both a Python API and a convenient command-line interface (CLI), allowing EspalomaCharge to be effortlessly integrated into popular MM and MD workflows such as the OpenForceField toolkit (Listing~\ref{lst:api-openff}) and Amber (Listing~\ref{lst:api-amber}).

\paragraph{One-hot embedding cannot generalize to rare or unseen elements}

One-hot element encoding is used in the architecture, making the model unable to perceive elemental similarities.
This would compromise per-node performance for rare elements and prevent the model to be applied on unseen elements.
Possible ways to mitigate this limitation include encoding the elemental physical properties as node input.

\paragraph{Future expansions of the training set could further mitigate errors}

As shown in Figure~\ref{fig:rmse_breakdown}, the generalization error is heavily dependent on the data abundance within the relevant stratification of the training set---bins containing more training data show higher accuracy.
Future work could aim to systematically identify underrepresented regions of chemical space and expand training datasets to reduce error for uncommon chemistries and exotic charge states, either with larger static training sets or using active learning techniques.

\paragraph{Multi-objective fitting could enhance generalizability}

Though EspalomaCharge produces accurate surrogate for AM1-BCC charges, these small errors in charges can translate to larger deviations in electrostatic potential (ESP) (Figure~\ref{fig:cdf}).
Since the function mapping charges (together with conformations) to ESPs are simple and differentiable, one can easily incorporate ESP as a target in the training process, using ESPs derived either from reference charges or (as in the original RESP~\cite{doi:10.1021/j100142a004}) to quantum chemical ESPs.
A multi-objective strategy that includes multiple targets (such as charges and ESPs), potentially with additional charge regularization terms (as in RESP~\cite{doi:10.1021/j100142a004}), could result in more generalizable models with lower ESP discrepancies.
Furthermore, similar observables can be incorporated into the training process to improve the utility of the model in modeling real condensed-phase systems. For instance, condensed phase properties such as densities or dielectric constants, other quantum chemical properties, or even experimentally measured binding free energies. 

\section{Funding}
Research reported in this publication was supported by the National Institute for General Medical Sciences of the National Institutes of Health under award numbers R01GM132386 and R01GM140090.
YW acknowledges funding from NIH grant R01GM132386 and the Sloan Kettering Institute.
JDC acknowledges funding from NIH grants R01GM132386 and R01GM140090.

\section{Disclaimer}
The content is solely the responsibility of the authors and does not necessarily represent the official views of the National Institutes of Health.

\section{Disclosures}
JDC is a current member of the Scientific Advisory Board of OpenEye Scientific Software, Redesign Science, Ventus Therapeutics, and Interline Therapeutics, and has equity interests in Redesign Science and Interline Therapeutics.
The Chodera laboratory receives or has received funding from multiple sources, including the National Institutes of Health, the National Science Foundation, the Parker Institute for Cancer Immunotherapy, Relay Therapeutics, Entasis Therapeutics, Silicon Therapeutics, EMD Serono (Merck KGaA), AstraZeneca, Vir Biotechnology, Bayer, XtalPi, Interline Therapeutics, the Molecular Sciences Software Institute, the Starr Cancer Consortium, the Open Force Field Consortium, Cycle for Survival, a Louis V.\ Gerstner Young Investigator Award, and the Sloan Kettering Institute.
A complete funding history for the Chodera lab can be found at \url{http://choderalab.org/funding}.

\section{Acknowledgments}
The authors would like to thank the Open Force Field consortium for providing constructive feedback, especially Christopher Bayly, OpenEye; David Mobley, UC Irvine; and Michael Gilson, UC San Diego.

\bibliography{main}

\appendix

\title{Appendix: EspalomaCharge: Machine learning-enabled ultra-fast partial charge assignment}
\maketitle
\section*{Detailed methods}
\label{sec:detailed}
\paragraph{Code availability.}
The Python code used to produce the results discussed in this paper is distributed open source under MIT license \url{https://github.com/choderalab/espaloma_charge}.
Core dependencies include PyTorch 1.12.1~\cite{paszke2017automatic}, Deep Graph Library 0.6.0~\cite{wang2019deep}, and the Open Force Field Toolkit 0.11.2~\cite{mobley2018open}.

\paragraph{Training dataset curation.}
The SPICE~\cite{https://doi.org/10.48550/arxiv.2209.10702} dataset is used as the training and in-distribution validation and test set for the model due to its thorough coverage of chemical space relevant to biomolecular simulations.
It consists of druglike small molecules selected from PubChem, short peptides, and fragments of biomolecules and biopolymers, and covers 15 elements (H, Li, C, N, O, F, Na, Mg, P, S, Cl, K, Ca, Br, I).
Protonation and tautomeric states have been enumerated for each molecule using the OpenEye toolkit.
After random shuffling (over the chemical space, protomeric and tautomeric states are kept in the same partition), 80\% of the dataset is used for training, 10\% used for validation (and model selection via early stopping), and 10\% for reporting out-of-sample test performance.

\paragraph{Out-of-distribution test dataset selection.}
To test the generalizability of EspalomaCharge, we select a series of out-of-distribution test datasets on which the discrepancy between charge methods are assessed (Table~\ref{tab:in_n_out}).
\begin{itemize}
    \item \textbf{FDA approved} dataset\footnote{Source: \url{https://github.com/openforcefield/qca-dataset-submission/tree/master/submissions/2019-09-08-fda-optimization-dataset-1}} contains FDA approved small molecules, filtered by size and element composition.
    \item \textbf{ZINC250K} dataset is a popular machine learning dataset first published in \citet{doi:10.1021/acscentsci.7b00572}, which randomly subsamples the original ZINC dataset~\cite{pmid15667143}.
    \item \textbf{FreeSolv} dataset~\cite{pmid15667143} contains small molecules whose hydration free energies have been experimentally measured. 
\end{itemize}

\paragraph{Neural network architecture and training}

Following the protocol specified in \citet{D2SC02739A}, we use GraphSAGE~\cite{hamilton2017inductive} as the GNN backbone and optimize the learning rate ($1e-2$ to $1e-5$), batch size ($16$ to $512$), and neural network width ($16$ to $512$) and depth ($2$ to $6$) via grid search on the validation set.
The input features of the atoms include the one-hot encoded element, as well as the hybridization, aromaticity, and (various sized-) ring membership, assigned using RDKit.
Note that the formal charges are not included as part of the features to avoid the time-consuming enumeration of resonance structure as in \citet{doi:10.1021/ci034148o}.
The hyperparameter search resulted in an optimal learning rate of $10^{-3}$ and L2 regularization with rate $10^{-4}$ with Adam optimizer~\cite{DBLP:journals/corr/KingmaB14} and batch size of $512$;
the neural networks are $4$ layers and $128$-unit wide.
All models were trained for 5000 epochs, and the model with optimal performance on the validation set was selected for characterization here.

\paragraph{Electrostatic potential (ESP) errors}

To calculate deviations between electrostatic potentials (ESP) on a surface, we first generated conformers using the OpenFF Toolkit 0.11.2. 
Conformer generation followed the Electrostatically Least-interacting Functional groups (ELF) approach. 
Initially, a maximum of 500 conformers was generated using RDKit with an RMS threshold of 0.05 \AA. A \textit{cis} conformation was enforced for carboxylic acid groups by rotating the protons in \textit{trans} carboxylic acids 180$^{\circ}$ around the C-O bond. The electrostatic energy of each conformer was calculated using MMFF94 charges~\cite{https://doi.org/10.1002/(SICI)1096-987X(199604)17:5/6<520::AID-JCC2>3.0.CO;2-W}. 
The 98\% conformers with the highest electrostatic energy were discarded. 
From the remaining 2\% conformers, we greedily selected up to 10 conformers that were most distinct from each other by RMS. Each conformer geometry was distinct by at least a heavy-atom RMS of 0.05 \AA~from each other.

For each conformer, we used OpenFF Recharge 0.4.0 to generate standard Merz-Singh-Kollman grids~\cite{https://doi.org/10.1002/jcc.540050204} around the molecule at a density of 1 point per \AA\textsuperscript{2}. 
We then calculate the root mean squared error (RMSE) between ESPs generated by each set of partial charges on the conformer grid. To compare the overall effect of different partial charges on the ESP, we average the RMSE between ESPs for each conformer.

\paragraph{Induced solvent potential from Poisson-Boltzmann model (ZAP)}

As a fast measure of how small differences in partial charges might impact interaction free energies, we computed the induced solvent potential on each atom using a fast Poisson-Boltzmann implicit solvation model implemented in OpenEye ZAP~\cite{grant2001smooth}.
The induced solvent potential reflects the potential induced by the polarization of the solvent, and was computed following recommended standard usage [\url{https://docs.eyesopen.com/toolkits/python/zaptk/thewayofzap.html}].

\paragraph{Hydration free energies in explicit solvent ($\Delta G_\mathrm{hyd}$)}

To compute hydration free energies for the FreeSolv dataset~\cite{mobley2018open} to quantify the impact of small differences in charges on experimentally-measurable free energies, we used a modified version of the protocol described in \citep{mobley2007comparison}.
Neutral molecules were solvated with TIP3P water~\cite{jorgensen1983comparison} in rectangular boxes with 14Å of padding, and assigned GAFF-2.11 parameters~\cite{wang2004development,wang2006automatic} using openmmforcefields~\cite{openmmforcefields}.
Hydration free energy calculations were computed by performing replica-exchange alchemical free energy calculations using a two-stage alchemical protocol in which charges were annihilated by linear scaling and Lennard-Jones interactions, and then subsequently annihilated using the Buetler softcore potential~\cite{beutler1994avoiding,pham2012optimal}.
Simulations employed particle mesh Ewald (PME)~\cite{essmann1995smooth} to treat long-range electrostatics and used mixed precision to ensure accuracy in energies and integration.
Integration was performed with the BAOAB Langevin integrator~\cite{leimkuhler2013rational,leimkuhler2013robust,leimkuhler2016efficient} using hydrogen masses of 3.8~amu to enable 4~fs timesteps to be taken while introducing minimal configuration space sampling error~\cite{fass2018quantifying}.
Calculations were carried out in gas phase at 298~K and in solvent at 1~atm using OpenMM~8~\cite{eastman2017openmm} and openmmtools~0.21.5~\cite{openmmtools}, and free energies were estimated with the multistate Bennett acceptance ration (MBAR)~\cite{shirts2008statistically} after automatic equilibration detection~\cite{chodera2016simple} and decorrelation.
Simulations were run for 1~ns/replica in each phase.
Code for reproducing these calculations can be found in \url{https://github.com/choderalab/espaloma_charge/tree/main/scripts/hydration-free-energies}.

\newpage
\begin{figure}
    \begin{center}
    \includegraphics[width=0.6\textwidth]{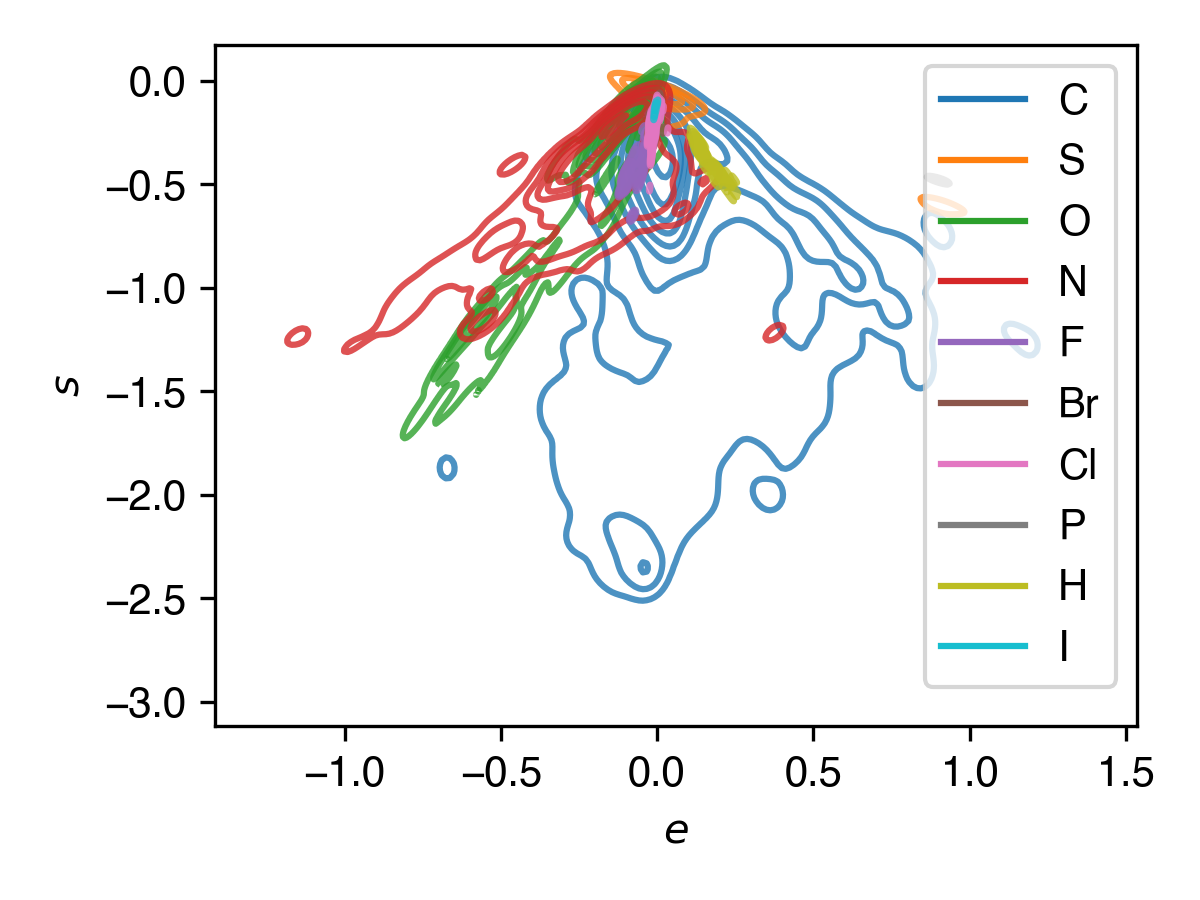}

    \end{center}

    \caption{
        \textbf{EspalomaCharge provides interpretable intermediate representations.}
        Kernel density estimate (KDE) plot of intermediate atomic electronegativity ($e$) and hardness ($s$) parameters used by the charge equilibration stage (Eq.~\ref{eq:charge-equilibration-solution}) to generate charges, stratified by element.
        While physical instances of these parameters are limited to being positive, in this model they are unconstrained in sign.
    }
    \label{fig:es}
\end{figure}

\begin{figure}
    \begin{center}
    \textbf{FreeSolv dataset}\\
    \includegraphics[width=0.8\textwidth]{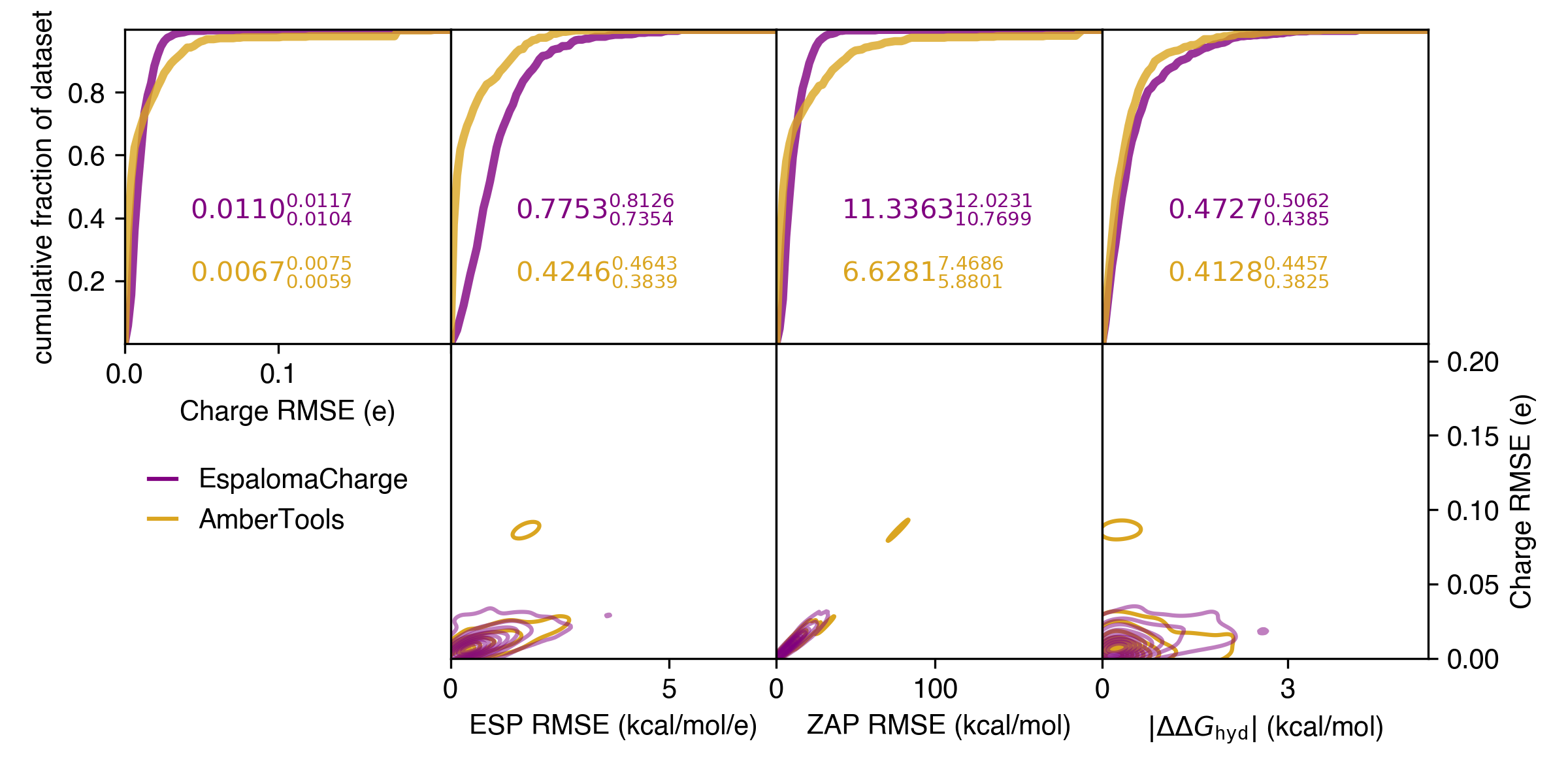}

    \textbf{FDA-approved dataset}\\
    \includegraphics[width=0.6\textwidth]{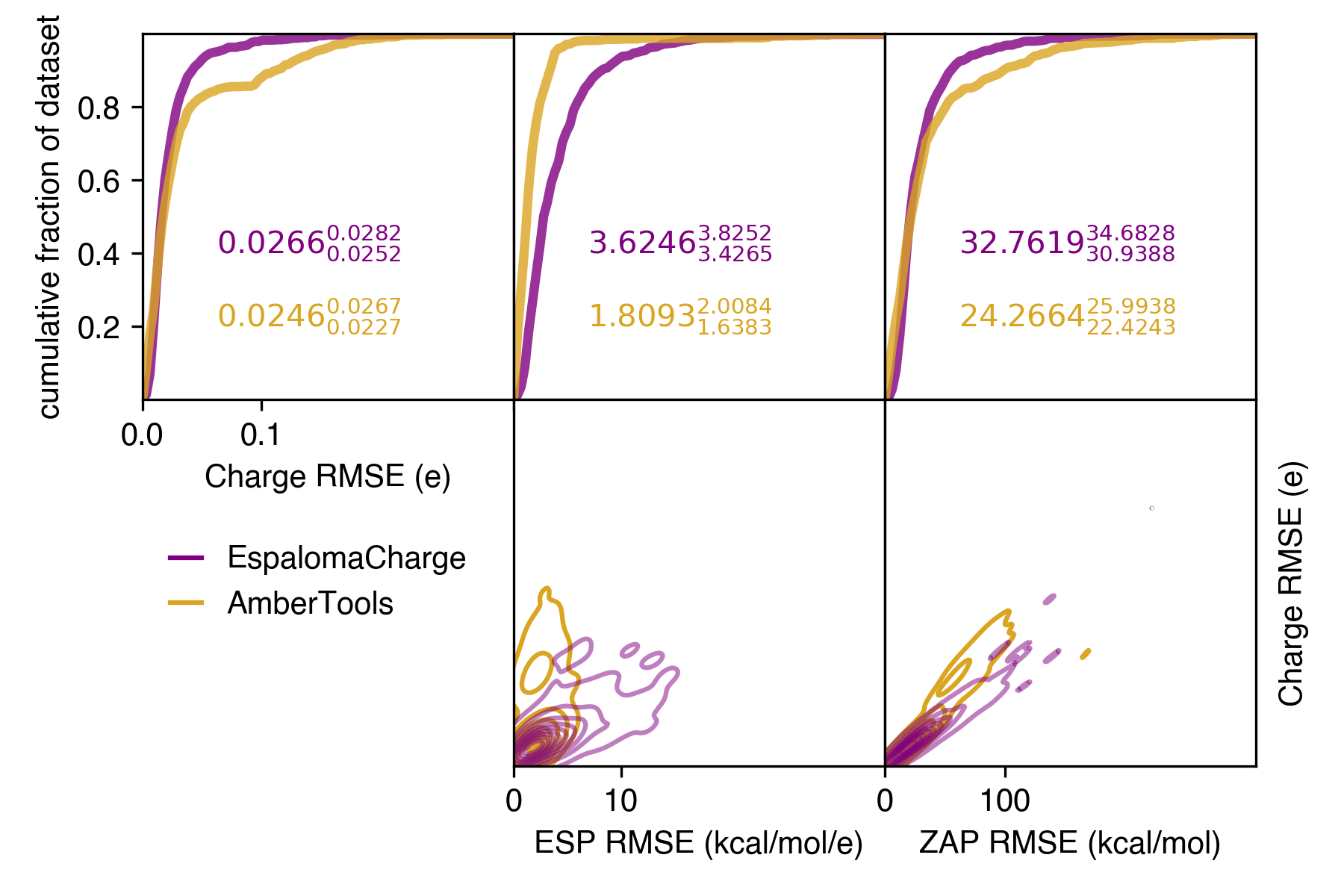}

    \end{center}

    \caption{
        \textbf{Comparison of discrepancies the EspalomaCharge and AmberTools \texttt{sqm} produce in computing various charge-dependent properties, with high-quality OpenEye AM1-BCC ELF10 charges taken here as ground truth.}
        The top row of each panel shows cumulative distribution functions (CDFs) of the deviations for each method, along with average (and 95\% bootstrapped confidence intervals) for EsplomaCharge (grapefruit) and AmberTools (amber).
        The bottom row of each panel shows the joint probability density functions (PDFs) of the deviations for each method in various properties along with the charge RMSE.
        Here, {\bf Charge RMSE (e)} denotes the root-mean squared (RMS) deviation of atomic charges for the molecule from OpenEye reference charges;
        {\bf ESP RMSE (kcal/mol/e)} denotes the RMS deviation of electrostatic potential on surface shells from OpenEye reference charges; 
        {\bf ZAP RMSE (kcal/mol)} denotes the RMS deviation in the induced solvent potential computed via the OpenEye ZAP fast Poisson-Boltzmann implicit solvent model solver~\cite{grant2001smooth} between the query charge model and the OpenEye reference charges;
        {$\bf \Delta \Delta G_\mathrm{hyd}$ (kcal/mol)} denotes the error in hydration free energies between the query charge model and the OpenEye reference charges.
        \emph{Top panel:} The FreeSolv dataset~\cite{mobley2018open} consists of 641 neutral small molecules with experimentally characterized hydration free energies.
        \emph{Bottom panel:} A subset of 1615 FDA-approved inhibitors (retrieved from ZINC~\cite{sterling2015zinc}, originally sourced from DrugBank~\cite{wishart2018drugbank}) with elements compatible with EspalomaCharge.        
    }
    \label{fig:cdf}
    
\end{figure}

\begin{figure}
    \centering
    \includegraphics[width=0.6\textwidth]{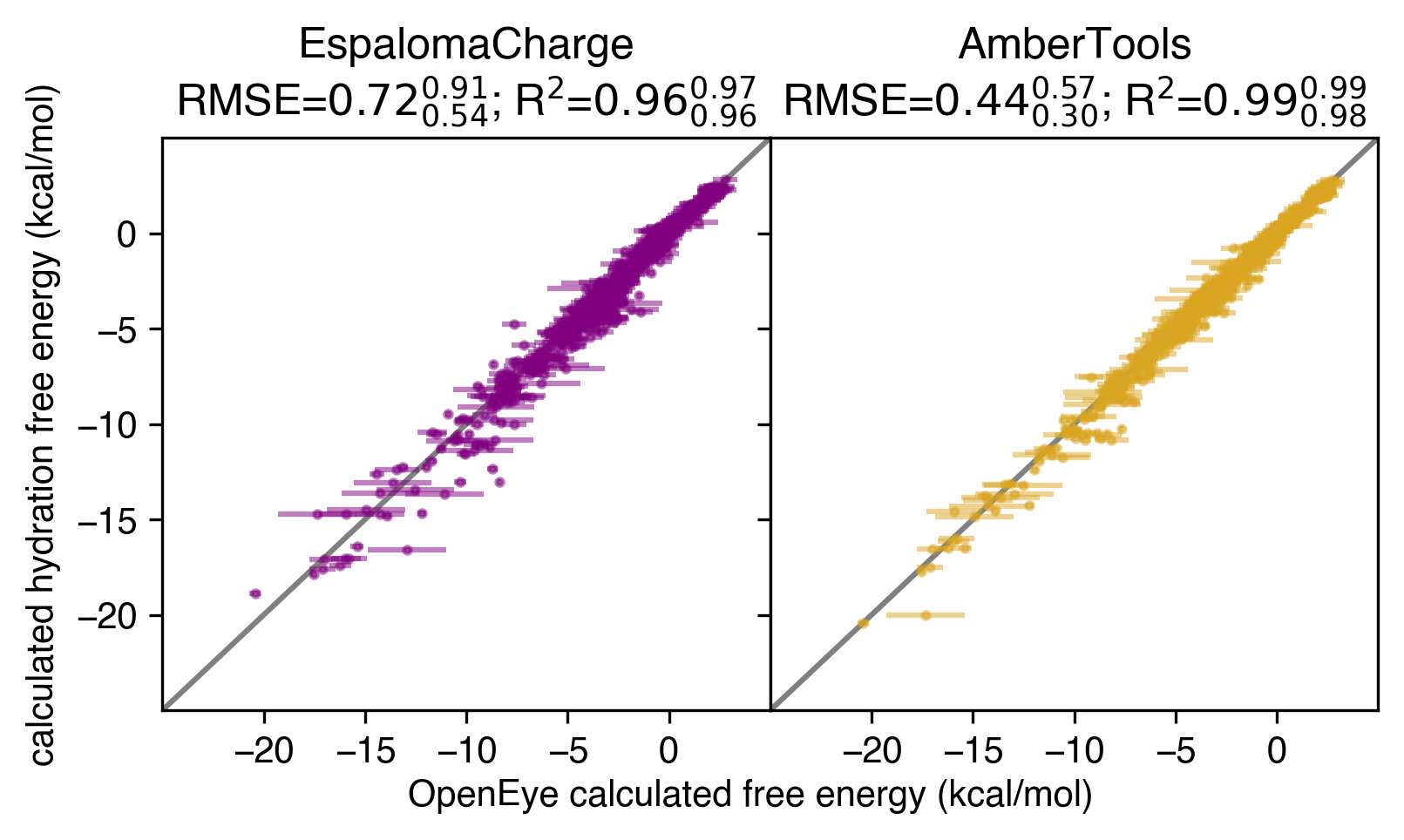}
    \caption{
    \textbf{EspalomaCharge introduces little error to explicit hydration free energy prediction.}
    Hydration free energy with EspalomaCharge- and AmberTools-calculated partial charges plotted against that generated with OpenEye-computed charge.
    }
    \label{fig:hydration_in}
\end{figure}

\end{document}